\documentclass[letterpaper, 10 pt, conference]{ieeeconf}
\usepackage{graphics} 
\usepackage{epsfig} 
\usepackage{mathptmx} 
\usepackage{times} 
\usepackage{amsmath} 
\usepackage{amssymb}  
\usepackage{bm} 
\usepackage{array}
\usepackage{subfigure}
\usepackage{cite}
\usepackage{url}
\usepackage{color}
\usepackage{threeparttable}
\graphicspath{{Figures/}}

\title{\LARGE \bf High-energy-density 3D-printed Composite Springs for \\Lightweight and Energy-efficient Compliant Robots}

\author{Amanda Sutrisno, Chase Mathews, and David J. Braun
\thanks{Amanda Sutrisno, Chase Mathews, and D. J. Braun are with the Advanced Robotics and Control Laboratory within the Center for Rehabilitation Engineering and Assistive Technology, Department of Mechanical Engineering, Vanderbilt University, Nashville, Tennessee 37235, USA.}
\thanks{E-mail: {\tt\small amanda.s.sutrisno@vanderbilt.edu}}
\thanks{E-mail: {\tt\small chase.w.mathews@vanderbilt.edu}}
\thanks{ E-mail: {\tt\small david.braun@vanderbilt.edu}}
}
\IEEEoverridecommandlockouts
\begin{document}
\maketitle
\begin{abstract}
Springs store mechanical energy similar to batteries storing electrical energy. However, conventional springs are heavy and store limited amounts of mechanical energy relative to batteries, i.e they have low mass-energy-density. Next-generation 3D printing technology could potentially enable manufacturing low cost lightweight springs with high energy storage capacity. Here we present a novel design of a high-energy-density 3D printed torsional spiral spring using structural optimization. By optimizing the internal structure of the spring we obtained a $45\%$ increase in the mass energy density, compared to a torsional spiral spring of uniform thickness. Our result suggests that optimally designed 3D printed springs could enable robots to recycle more mechanical energy per unit mass, potentially reducing the energy required to control robots.
\end{abstract}
\section{Introduction}

Springs are standard parts which can store and release mechanical energy. They are used in modern robotics applications to recycle energy in cyclic tasks such as throwing objects \cite{Braun2012, Braun2013}, lifting \cite{ Lau2018}, jumping \cite{ Kovac2009, Sutrisno2019}, walking \cite{Collins2015, Zhang2022}, running \cite{Thompson1990, Hurst2004, Nasiri2018, Sutrisno2020}, and to make compliant actuators \cite{Pratt1995,Braun2019,Braun2019a,Mathews2021,Kim2021}. However, the utility of a spring is limited by the mass-energy-density of a spring \cite{Stucheli2016}.
Commercially available steel spring designs possess an energy density around $10-150$~J/kg \cite{Carpino2012, Rossi2015, Georgiev2017}, which is far below the theoretical maximum energy density of steel, $1400-1700$~J/kg. The discrepancy between the mass-energy-density of current springs and the theoretical upper limit shows that spring designs have significant room to improve.

Optimization can be used to vary spring parameters to minimize or maximize a desired objective function, for example, minimize the weight of the spring while storing the same amount of energy \cite{ Berger2017, Zhu2020, Vijay2022}. 
Prior works \cite{Paredes2001, Scarcia2016} to optimize spring designs often took into account conventional manufacturing constraints, such as optimizing wire thickness or coil radius of a coil spring, which is made from twisting a wire of uniform thickness around a cylinder of uniform radius. However, with the rise of additive manufacturing, the entire structure of the spring, and not just a few dimensions of a predefined spring shape, can be optimized. Perhaps this additional manufacturing freedom in generating more complex structures could enable the design of higher energy density springs.

\begin{figure}
	\includegraphics[width=\linewidth]{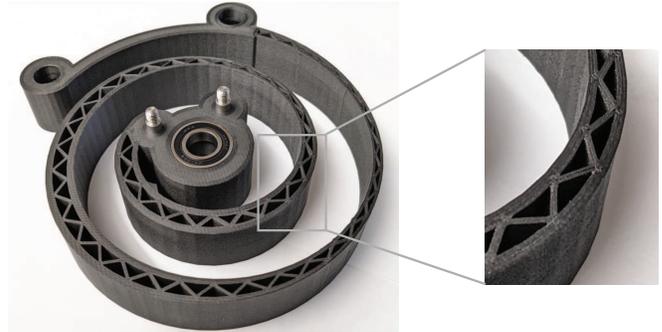}
	\caption{Spiral spring optimized for high energy density. }
	\label{Fig:spring}
\end{figure}

In this paper, we show that optimizing the internal structure of a 3D printed torsional spiral spring can be used to significantly increase the mass-energy-density of the spring \cite{Guo2022}. We have experimentally obtained a $45$\% increase in energy density when optimizing the thickness and internal structure of a spiral spring, compared to another spiral spring of similar geometry, uniform thickness, and solid infill. The new design was found by first modeling the spring as an Euler-Bernoulli beam with large deformations, then optimizing the spring along its length to thicken sections with higher loading and conversely make thinner sections with minimal loading. We then improve energy density further by simulating the spring using Finite Element Analysis software (ANSYS\textsuperscript{\textregistered} Mechanical\textsuperscript{TM}), to remove material near the neutral axis to make trusses. 

Increasing the energy density of springs could enable passive mechanisms to become viable in applications with severely restricted limits on mass, for example, autonomous robots and devices built for human augmentation. A typical running shoe does not store significant amounts of energy, but results in a human wearer being able to run faster due to their negligible weight, compared to spring-leg exoskeletons that typically weigh significantly more than running shoes. 
Minimizing the weight of springs could result in increased mobility and more efficient motion in both robots and humans augmented using wearable spring exoskeletons.

\begin{figure*}
	\begin{center}
		\includegraphics[width=1\linewidth]{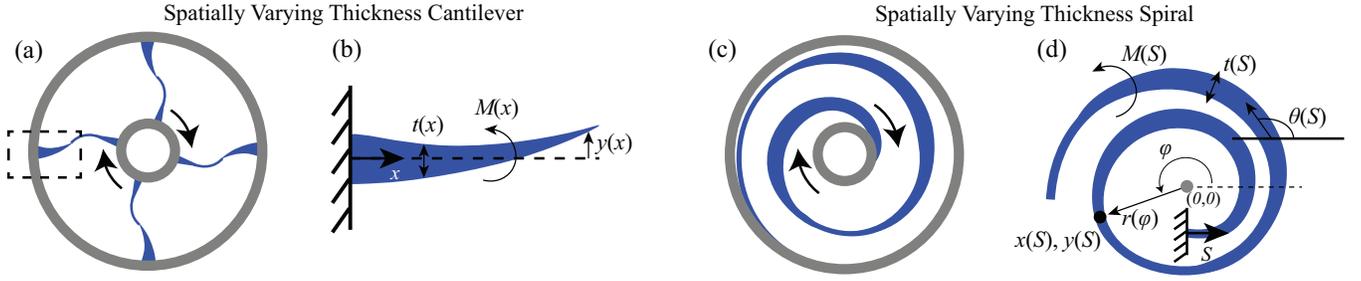}
	\end{center}
	\caption{Optimization of torsion spring designs. (a) Example of torsion spring composed of cantilevers. (b) Cantilever with spatially varying thickness. (c) Torsion spring composed of a spiral. (d) Model of a spiral with spatially varying thickness.}
	\label{Fig:torsion_springs}
\end{figure*}

\section{Prediction using a Simple Model}
The main idea to design a high energy density spring is to remove material from sections of the spring which store little energy under an applied load. 


%
Let us first define the potential energy of a bending beam
\begin{align}
	U = \frac{1}{2}\int_{0}^{L} \frac{M(x)^2}{E I}dx,
\end{align}
where $M(x)$ is the moment along the beam, $L$ is the total length of the beam, $x$ is the position along the length of the beam, $E$ is the Young's modulus of the beam material, and $I$ is the second moment of inertia of the beam cross-section. 
Furthermore, the mass of the beam of length $x$ is $m(x)=\rho w tx$, where $\rho$ is the density of the material, $w$ is the width of the beam, and $t$ is the thickness of the beam, while the mass-energy-density of the beam is given by
\begin{align}
	\frac{dU}{dm} = \frac{dU}{dx} \bigg(\frac{dm}{dx} \bigg)^{-1}
	= \frac{1}{2} \frac{M(x)^2}{E I} \frac{1}{\rho w t} .
	\label{eq_dUdm_simple}
\end{align}

Let us now look at the simple problem of a cantilevered flat beam of uniform thickness under a transverse load at the free end, shown in Fig.~\ref{Fig:torsion_springs}ab. For this problem, the moment along the beam is given by
\begin{align}
M(x)=F(L - x), \label{moment}
\end{align}
where $F$ is the vertical force at the free end of the beam.

We can see from (\ref{moment}) that the material is maximally loaded near the cantilevered end $x=0$, and decreases in energy stored per unit length until zero at the free end (\ref{eq_dUdm_simple}). We can also see that the uniform thickness cantilevered beam is inefficient because the amount of material throughout the beam is uniform despite the energy stored being non-uniform. In this simple example, it is straightforward to increase the material near the cantilevered side and decrease the material near the free end of the beam. Specifically, assuming a variable thickness beam $t=t(x)$, and using $I=w t(x)^3/12$, we can vary the thickness of the beam along the length of the beam to obtain:
%
\begin{align}
	\frac{dU}{dm} = \frac{6 F^2(L-x)^2}{ E \rho w^2 t(x)^4}. \label{eq_dUdm_x}
\end{align}

For every material, the maximum energy density for a pure bending load is defined by $dU_{\max}/dm = \sigma_{\max}^2/(6 E \rho)$, where $\sigma_{\max}$ is the yield tensile stress of the material. Therefore, to obtain the maximum energy storage density in the cantilevered beam, the energy density must be equal to $dU_{\max}/dm$, which we may substitute into (\ref{eq_dUdm_x}) to find the optimal thickness of the beam:
\begin{align}
	t(x) &=  \sqrt{\frac{6 F(L-x)}{w \sigma_{\max}}}.
\end{align}
 
This example shows that 
it is possible to maximize the energy density in a beam by varying the thickness along the length of the beam.

We can use a similar approach to make a spiral for large deformation torsional motion by modeling a spiral spring using large deformation Euler-Bernoulli beam theory. 
Torsion springs are useful in robotics because they can be attached to rotational joints, including human joints to augment motion\cite{ Li2019, Mathews2021}.
This can be done by creating two concentric circles of different radii connected by a spiral spring, in which the circles rotate in opposite directions of each other to store energy. Despite additional modeling complexity, a spiral spring can be beneficial due to its larger maximum deflection comparable to its length, since a cantilever is only capable of small deformations relative to its length.

In the next section we shall show how the internal structure of a spiral spring can be optimized to maximize the energy density of the spring.

\section{Design Modeling}
In this section we first present a model (Section~\ref{section_model}) for the spiral spring, which we shall use to describe the optimization problem (Section~\ref{section_opt}) and the resulting optimal shape obtained through simulations (Section~\ref{section_theory_pred}).

\label{section_model}
In order to model a spiral spring shown in Fig.~\ref{Fig:torsion_springs}cd, we use the nonlinear large-deformation version of the Euler-Bernoulli beam equations\cite{Georgiev2017}:
\begin{align}
	\frac{dM}{dS} &= - V \cos(\theta) + H \sin(\theta), \quad
	\frac{d\theta}{dS} = \frac{d\theta_0}{dS} - \frac{M}{E I(S)}, \nonumber \\
	\frac{dx}{dS} &= \cos(\theta), \quad \frac{dy}{dS} = \sin(\theta) \label{eq_EB_nonlin}
\end{align}
where $S\in[0,L]$ is the arc length along the beam, $M(S)$ is the bending moment, $V$ is the vertical force, $H$ is the horizontal force, $\theta(S)$ is the angle the beam makes with the $x$-axis, $\theta_0(S)$ is the initial curvature of the beam, while $x(S),y(S)$ define the axis of the beam, $E$ is the Young's modulus of the beam material, and $I(S)$ is the second moment of inertia of the beam cross-section.  

We would like to model the spring using an Archimedean spiral. 
Geometrically, the center of the beam in the shape of an Archimedean spiral can be defined in polar coordinates,
\begin{align}
	r = r_0  + \frac{\phi}{\phi_{\max}} \Delta r,\;\;\;
	x = r(\phi)\cos(\phi),\;\;\; y = r(\phi) \sin(\phi) \label{eq_arch_spiral}
\end{align}
where $\phi \in [0, \phi_{\max}]$ is the polar angle of the location of the beam (to distinguish from $\theta$, the angle of the tangent of the beam), and $r$ is the distance of the beam from the origin. The shape of the Archimedean spiral is encoded in $d\theta_0(S)/dS=(d\theta_0/d\phi) (d\phi/dS)$ and $\theta_0(0)$, which can be derived from (\ref{eq_arch_spiral}); due to the analytical complexity, the derivation is not presented here. 

\begin{figure*}
	\includegraphics[width=\linewidth]{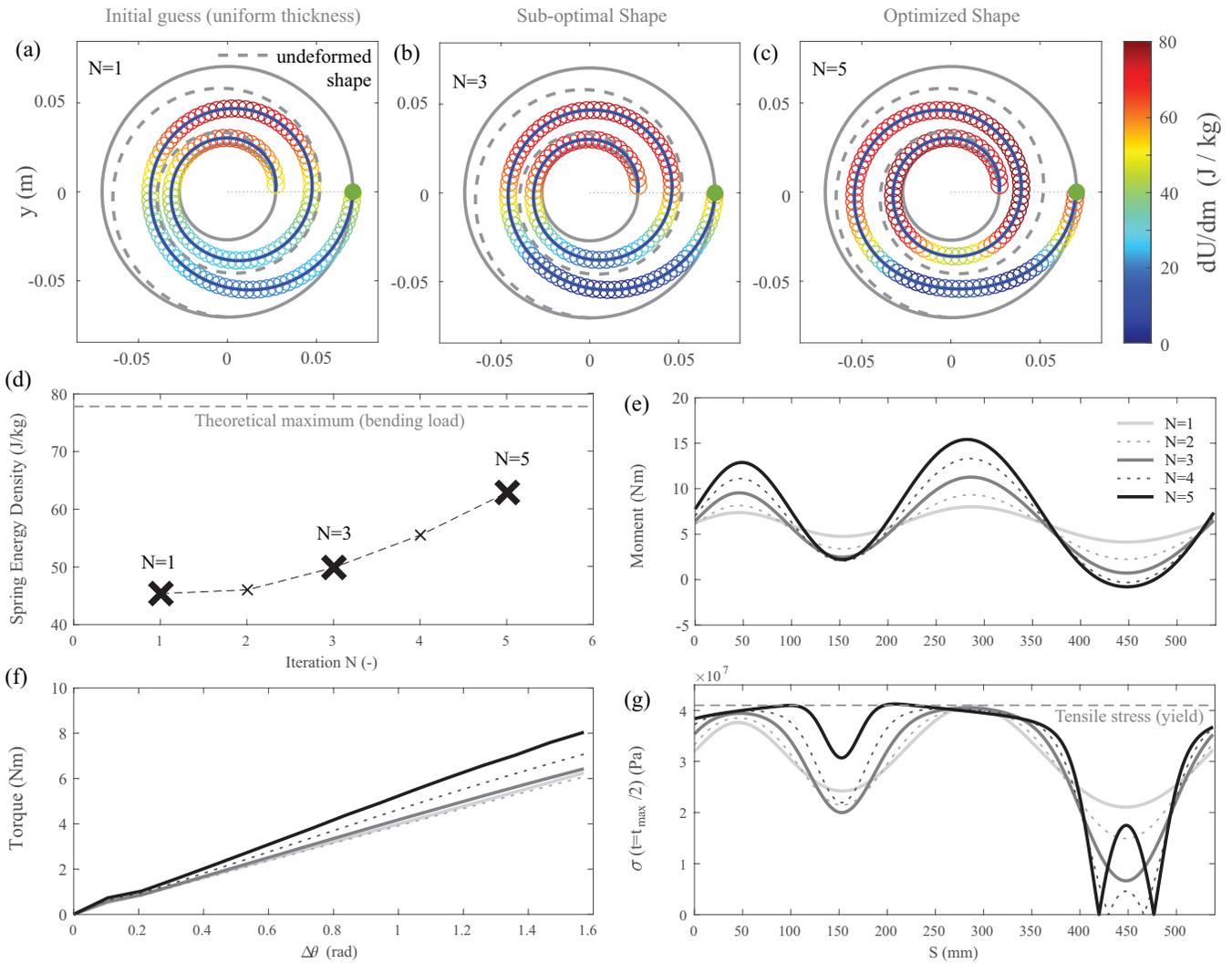}
	\caption{Optimized model of a spiral spring. (a) Energy density plot of a spring with a uniform thickness. (b) Energy density plot of a spring with spatial varying thickness. (c) Energy density plot of the spring with optimal varying thickness. (d) Total energy density over successive iterations. (e) Moment along beam length. (f) Torque-deflection relation. (g) Maximum stress along beam length.}
	\label{Fig:spring_opt}
\end{figure*}

Finally, we use the following boundary conditions for subjecting the outer end of the spiral to a twist of $\Delta \phi$ angle:
\begin{align}
	&x(0) = r_0,\;\;
	y(0) = 0,\;\;
	\theta(0) = \tan^{-1} \bigg(\frac{dy}{d\phi} \bigg(\frac{dx}{d\phi} \bigg)^{-1} \bigg) \bigg|_{\phi = 0}, \nonumber \\
	&x(S_{\max}) = (r_0 + \Delta r) \cos(\phi_{\max} + \Delta \phi),\nonumber\\
	&y(S_{\max}) = (r_0 + \Delta r) \sin(\phi_{\max} + \Delta \phi),\nonumber\\ 
	&\theta(S_{\max}) = \theta_0(S_{\max}) + \Delta \phi. \label{eq_bc}
\end{align}

In equations (\ref{eq_EB_nonlin}), the unknown quantities $M(0), V, H$ are chosen to meet the last three conditions in (\ref{eq_bc}).
\section{Optimization}
\label{section_opt}
Given the Archimedean spiral geometry described by $d\theta_0/dS$, we would like to optimize the thickness of the spiral $t(S)$ in order to maximize the energy density. This can be formulated as an optimization problem,
\begin{align}
	\max_{t(S)} U[t(S)] 
	 \label{eq_obj}
\end{align}
subject to the following constraints
\begin{align}
\frac{dU}{dm} \leq \frac{\sigma_{\max}^2}{6 E \rho}\;\;\;\; \text{and} \;\;\;\; t(S) \geq t_{\min}>0, \label{eq_constr}
\end{align}
where (\ref{eq_constr}) indicates that the energy density at any particular point cannot exceed the fracture limits of the material, and the thickness cannot be smaller than the limit set by 3D printing constraints.

Unlike the simple example of a cantilevered beam with small deformations described in the previous section, this problem cannot be solved analytically because the moment $M(S)$ depends on the thickness $t(S)$, and can only be computed numerically by solving the nonlinear boundary value problem (\ref{eq_EB_nonlin}) and (\ref{eq_bc}). Consequently, the optimization has to be done numerically. 

In order to solve the optimization problem defined in (\ref{eq_obj})-(\ref{eq_constr}), we have implemented a simple iterative method that gradually optimizes the thickness of the spring 
using the recursive formula
\begin{align}
	t_{n+1}(S) = c_1 t_{n}(S) e^{-c_2(1 - \frac{dU(S)}{dm}/ \frac{dU_{\max}}{dm})}, \label{eq_opt_algo}
\end{align} 
where $n$ denotes the number of iterations, $dU_{\max}/dm$ is the maximum of the mass-energy-density defined in (\ref{eq_constr}), $c_1$ is a positive constant which is used to scale the thickness of the beam to ensure that the first condition in (\ref{eq_constr}) is satisfied, while $c_2$ is a positive constant which controls how fast material is re-distributed along the length of the beam. The recursive formula (\ref{eq_opt_algo}) cannot make the spring thickness negative and stops removing material when $t(S) = t_{\min}$ according to (\ref{eq_constr}).

\begin{figure*}
	\includegraphics[width=\linewidth]{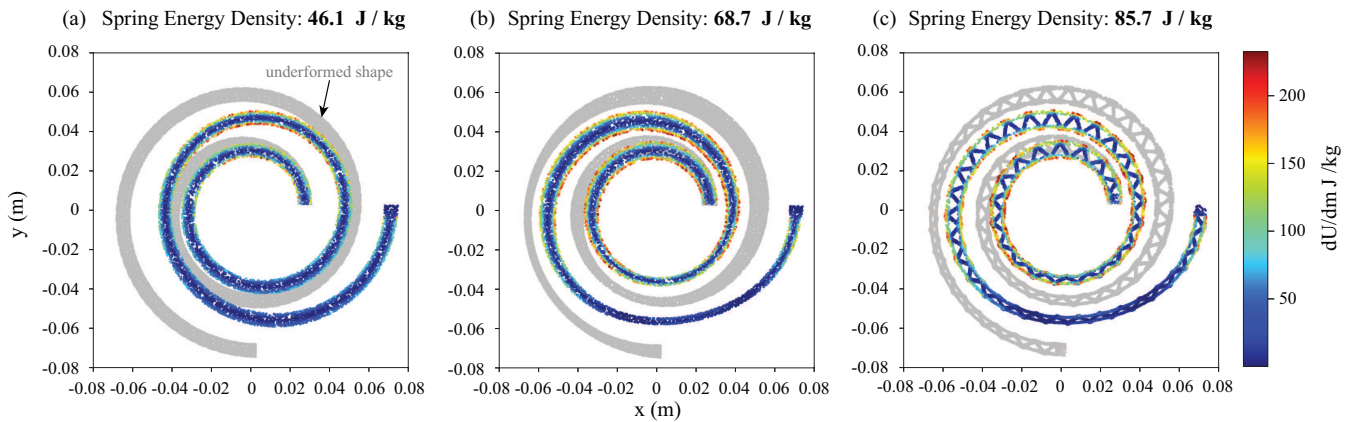}
	\caption{Energy density plots of torsion springs computed using ANSYS\textsuperscript{\textregistered} Mechanical\textsuperscript{TM}. (a) Uniform thickness spiral spring. (b) Optimized spatially varying thickness spiral spring. (c) Optimized spatially varying thickness spiral spring with material cut near neutral axis. The gray shapes show the un-deformed springs.}
	\label{Fig:ansys}
\end{figure*}

\section{Theoretical Prediction}
\label{section_theory_pred}
In this section we summarize the prediction obtained by structural optimization of a spiral torsional spring. 
 
 \subsection{Optimization}
We use the method summarized in (\ref{eq_opt_algo}) to solve the optimization problem defined by (\ref{eq_obj})-(\ref{eq_constr}).
The optimization was done using the dimensions and material properties in Table~\ref{Table:prop} (we assumed the spring is made of the Onyx 3D printing material for the Mark Forged Mark II 3D printer). The results of the optimization is shown in Fig.~\ref{Fig:spring_opt}.

\renewcommand{\arraystretch}{1.2}
\begin{table}
	\caption{Spiral Torsion Spring Dimensions and material properties.}
	\label{Table:prop} 
	\normalsize
	\begin{tabular}{  m{1cm}  m{4cm} m{1cm}  m{1cm} }
		\hline
		Symbol & Description & Value & Units  \\ \hline
		$r_0$ & inner radius & 27 &  mm \\ 
		$r_1$ & outer radius & 70.5 & mm \\ 
		$\phi_{\max}$ & final polar angle & $3.5 \pi$ & rad \\ 
		$w$ & spring width & 20 & mm \\ 
		$t_0$ & initial uniform thickness & 7 & mm \\ 
		$E$ & Young's modulus & $3.0$ & GPa \\ 
		$\rho$ & density & 1200 & kg/$m^3$ \\ 
		$\sigma_{\max}$ & yield strength & 41 & MPa \\ 
		$\frac{\sigma_{\max}^2}{6 \rho E}$ & maximum energy density in pure bending& 78.0 & J/kg \\ 
		\hline
	\end{tabular}
\end{table}

The initial guess of a uniform thickness spiral has a total energy density of $45$~J/kg, which is $60$\% of the theoretical maximum energy density under a pure bending load (assumes all parts of the spiral are under maximum bending load). After four iterations, the energy density increased to $63$~J/kg, which is $79$\% of the theoretical maximum, see Fig.~\ref{Fig:spring_opt}d. According to Fig.~\ref{Fig:spring_opt}abc, the successive optimizations result in a more even distribution of energy density along the length of the beam, whereby less than half of the length of the spring in the initial guess (a) is near the maximum possible load, while the optimized design (c) has over $80$\% of the spring length loaded to the maximum stress, see Fig.~\ref{Fig:spring_opt}g. 

The optimization was done by successively decreasing the thickness in areas with low moment load and increasing the thickness in zones with a high amount of moment load, see Fig.~\ref{Fig:spring_opt}e. Although the effect of adding variable thickness to the spring has the unintended effect of increasing the variability in the moment, there is more material in areas with high moment, and less material in areas with low moment which results in a higher overall energy density.

\subsection{Removing Material around the Neutral Axis}
Some prior works assumed the cross-section of the spring is a solid rectangle \cite{Carpino2012,Scarcia2016}, whilst others have cut away material near the center of the spring and the neutral axis to obtain better energy density \cite{Georgiev2017, Vijay2022}. We used ANSYS\textsuperscript{\textregistered} Mechanical\textsuperscript{TM} to show that cutting away material near the center of the spring increases the energy density of the spring. In Fig.~\ref{Fig:ansys}, we compared a solid spiral of uniform thickness (a), a solid spiral of optimized spatially varying thickness (b), and a spiral of spatially varying thickness with material cut out of the middle to form trusses (c).

According to Fig.~\ref{Fig:ansys}ab, the total mass energy density for the solid springs is similar to the energy density computed by the Euler-Bernoulli equation in (\ref{eq_EB_nonlin}), $46$~J/kg versus $45$~J/kg for the uniform thickness spiral, and $68.7$~J/kg versus $63$~J/kg for the variable thickness spiral. According to Fig.~\ref{Fig:ansys}ab, the energy stored is mainly concentrated in the outer walls of the spiral, with almost no energy stored in the middle, near the neutral axis of bending. This feature can be exploited to further increase the mass energy density of the spring by hollowing out the variable thickness spiral with trusses. Figure~\ref{Fig:ansys}c shows the variable thickness hollow spring which has the mass energy density of $85.7$~J/kg. This theoretically predicted mass energy density is a $25$\% increase over a solid shape with optimized thickness, and an $86$\% increase over the solid spiral with uniform thickness. 

\section{Experimental Evaluation}
In this section we describe the experimental setup used to characterize the 3D printed springs, the procedure used to measure the torque-deflection data, and the results of the characterization.
\subsection{Setup}
\begin{figure}
	\includegraphics[width=\linewidth]{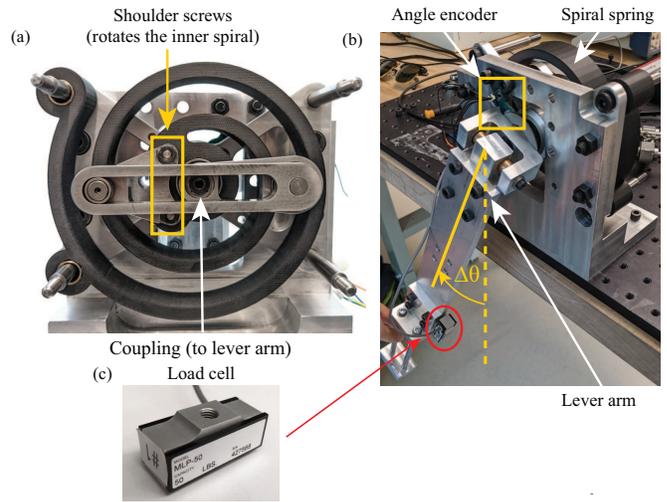}
	\caption{Experimental setup. (a,b) The spring was mounted into a variable stiffness mechanism presented in \cite{Mathews2021}, operated in fixed stiffness mode. (c) A load cell used to measure the torque generated by the spring.}
	\label{Fig:setup}
\end{figure}

The setup is comprised of a freely rotating lever-arm which couples its rotation to the inner section of the spiral spring, in which the outer parts of the spiral are constrained to be stationary, see Fig.~\ref{Fig:setup}ab. The lever arm is $0.24$~m long, and has a load cell (Transducer Techniques MLP-50) attached to the outer end of the lever, see Fig.~\ref{Fig:setup}c, which is capable of recording an applied force. The lever arm is also instrumented with a magnetic encoder (AMS AS5304), see Fig.~\ref{Fig:setup}b, used to measure the rotation of the lever arm $\Delta \theta$.

\subsection{Procedure}
We characterize spiral springs by measuring the torque-deflection profile of the spring under a quasi-static load. We started recording the force on the load cell and the angular deflection at the equilibrium position of the spring, when no forces are applied to the lever arm. Then we slowly deflected the angle of the lever arm by pushing only on the load cell mounted to the lever arm until the spring was deflected 90 degrees. Finally, we slowly return the lever arm to the equilibrium position. The torque exerted on the spring is obtained by multiplying the force on the load-cell by the length of the lever arm $0.24$~m.

\begin{figure*}
	\includegraphics[width=\linewidth]{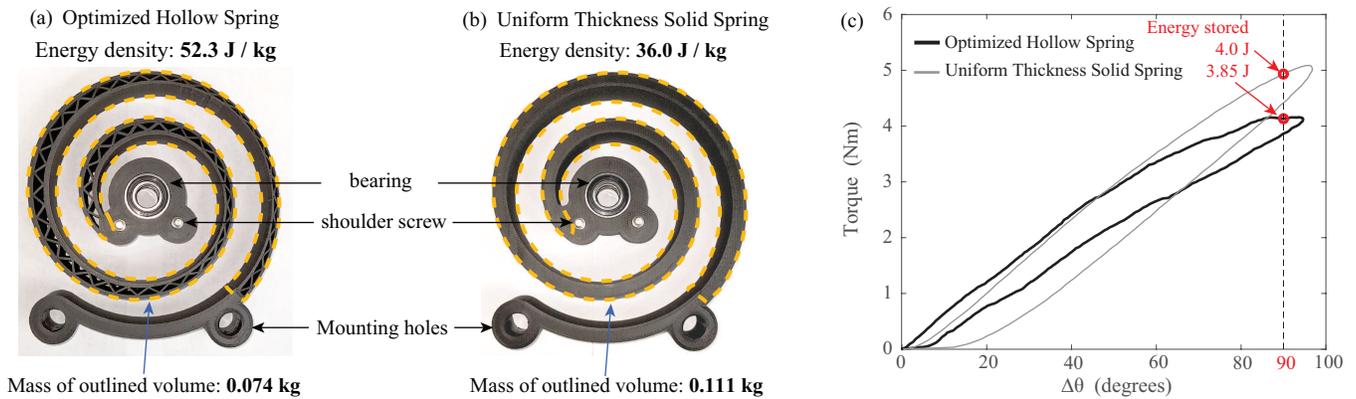}
	\caption{3D printed springs. (a) Optimized design. (b) Spring of uniform thickness and solid infill. Outlined area indicates the parts of the spring used in mass calculations. (c) Torque-deflection data, where energy stored is defined at a 90~degree deflection. Both springs show similar energy stored, but the mass of the optimized spring is $33\%$ lower. }
	\label{Fig:experiment}
\end{figure*}

\subsection{Results}
We have 3D printed (Mark-Forged Mark II, Onyx material) and experimentally characterized the final optimized spring, see Fig.~\ref{Fig:ansys}c. As a control experiment, we have 3D printed a solid in-fill spiral of uniform thickness with the same dimensions as the spring simulated in Fig.~\ref{Fig:ansys}a. Figure~\ref{Fig:experiment}ab shows the 3D printed springs, and Fig.~\ref{Fig:experiment}c shows the resulting torque-deflection plots.

The optimized 3D printed spring stored $3.85$~J energy at 90 degrees deflection, and had a mass of $0.140$~kg, resulting in a mass-energy-density of $27.5$~J/kg, while the un-optimized uniform thickness solid spiral resulted in $4$\% more energy stored, $4.0$~J at 90 degrees, and a larger total mass of $0.176$~kg, resulting in a lower energy density of $22.7$~J/kg. 

If we exclude the material used to mount the springs to the experimental setup and only count the volumes of the 3D printed spring which undergo deformation (outlined areas in Fig.~\ref{Fig:experiment}), then the optimized spring weighs only $0.074$~kg, and has a mass energy density of $52.3$~J/kg, whilst the uniform thickness spring weighs $0.11$~kg and has a mass energy density of $36.0$~J/kg, representing a $45$\% total increase in mass energy density of the optimized spring versus the uniform spring. 

\section{Discussion and Conclusion}
In this paper we have taken a classic torsion spiral spring design, modeled the spring using nonlinear Euler-Bernoulli beam equations, and optimized the spring for mass-energy-density. Bellow we discuss the challenges in designing springs with high mass-energy-density, and the implications of using such springs in robotics.

The simulation and optimization method presented in this paper resulted in a theoretical $86$\% increase in the spring energy density when combining thickness optimization \cite{Ahmed2014} and hollowing out material near the neutral axis of the spring \cite{Georgiev2017,Vijay2022}. 
However, optimizing the geometry of the spiral could potentially allow for improvements. For example, one may optimize the number of en-circlements of the spiral, or the rate of change of the spring radius with respect to polar angle. Furthermore, while we predicted $85.7$~J/kg mass energy density for the optimized spring, this prediction is less than $40$\% of the theoretical limit $\sigma_{\max}^2/(2 E \rho) = 233$~J/kg of a material homogeneously loaded to yield stress. This implies there may be better designs that do not even resemble spiral springs.

Spring-like mechanisms could reduce the cost of robot walking by 40-50\%\cite{Mazumdar2015}, and in theory enable the design of passive robots, without batteries or motors, capable of self-sustained locomotion \cite{Thompson1990}. However, current spring designs do not possess sufficient mass energy density to augment physically demanding tasks \cite{Sutrisno2020}. 
The development of high energy density springs could enable passive mechanisms to become more feasible, reducing the energy cost of locomotion to increase robot and human mobility.
\bibliographystyle{ieeetr}
\bibliography{citations}
\end{document}